\documentclass[10pt,twocolumn,letterpaper]{article}

\usepackage{iccv}
\usepackage{times}
\usepackage{epsfig}
\usepackage{graphicx}
\usepackage{amsmath}
\usepackage{amssymb}
\usepackage{algorithm}
\usepackage{algorithmic}

\usepackage[breaklinks=true,bookmarks=false]{hyperref}

\iccvfinalcopy 

\def\iccvPaperID{****} 

\begin{document}

\title{Are all training examples equally valuable?}

\author{Agata Lapedriza\\
Massachusetts Institute of Technology\\
Universitat Oberta de Catalunya \\
Computer Vision Center \\
{\tt\small agata@mit.edu}
\and
Hamed Pirsiavash \\
Massachusetts Institute of Technology\\
{\tt\small hpirsiav@mit.edu}
\and
Zoya Bylinskii\\
Massachusetts Institute of Technology\\
{\tt\small zoya@mit.edu}
\and
Antonio Torralba \\
Massachusetts Institute of Technology\\
{\tt\small torralba@mit.edu}
}

\maketitle

\begin{abstract}

When learning a new concept, not all training examples may prove equally useful for training: some may have higher or lower training value than others. The goal of this paper is to bring to the attention of the vision community the following considerations: (1) some examples are better than others for training detectors or classifiers, and (2) in the presence of better examples, some examples may negatively impact performance and removing them may be beneficial. In this paper, we propose an approach for measuring the training value of an example, and use it for ranking and greedily sorting examples.  We test our methods on different vision tasks, models, datasets and classifiers. Our experiments show that the performance of current state-of-the-art detectors and classifiers can be improved when training on a subset, rather than the whole training set. 

\end{abstract}

\section{Introduction}

When developing an object detection system, the first challenge involves choosing a training dataset, for instance out of the currently popular datasets~\cite{PASCAL,SUN,IMAGENET}. Standard practice is then to treat all the training examples equally, by feeding the full training set into the learning algorithm. 
Indeed, as more training data becomes available, detector performance will tend to increase. This is illustrated in Fig.~\ref{fig:multiplots}\textcolor{red}{.b} (black curve) where Average Performance (AP) is plotted as a function of the number of training examples.

\begin{figure}
\centering
\includegraphics[width=1\linewidth]{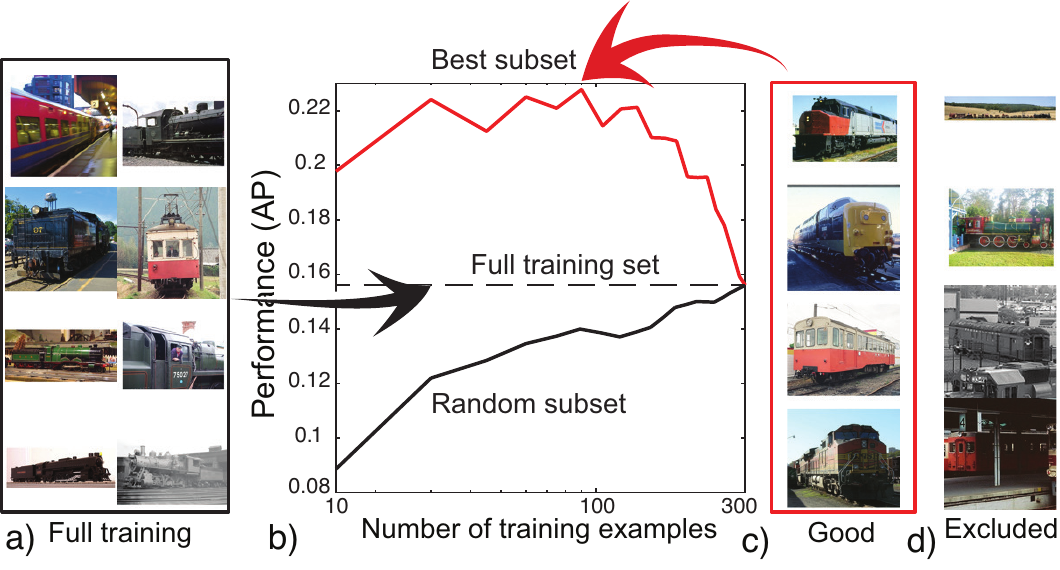}
\caption{a) Training examples from the {\em train} class from PASCAL 2007. (b) If we take larger subsets of the training set, performance on the test set grow (black curve). When we train the classifier with increasing set sizes, but consider examples ranked from greater to lower training value, we observe a very different trend (red curve). Performance first grows, reaches a maximum and then drops, converging on the performance achieved when training with the full training set. We depict some examples with (c) higher training value and (d) lower training value.}
\label{fig:multiplots}
\vspace{-.1in}
\end{figure}

However, when learning a new concept, not all training examples may prove equally useful for training.
For instance, mislabeled or inaccurately demarcated examples may actually hurt the performance of the classifier, and thus be less valuable for training. More interestingly, even if all the examples are correctly segmented and labeled, not all of them will be equally valuable in training a model for a given concept (under the limitations of that particular model). For instance, if a model can not handle occlusions, then a matching algorithm based on this model could be hurt by partially occluded and truncated training examples. 
In such a case, adding more training examples with large amounts of occlusion may actually hurt rather than improve the performance of the final classifier. 

The red curve in Fig.~\ref{fig:multiplots}\textcolor{red}{.b} illustrates how performance changes if the same examples as used for the black curve are now sorted according to training value. This performance plot demonstrates that if we add training examples by sorting them first (Fig.~\ref{fig:multiplots}\textcolor{red}{.c,d}), then performance increases faster with increasing training set size than if we were to add training examples in a random order. More interestingly, the performance achieved by using only a subset of the training set is actually higher than when using the entire training set. This can be seen in the plot, where the maximal performance achieved by the best subset climbs above the performance achieved with the full training set. 

This is not just a phenomenon of this example, but rather we show in the paper that this often occurs when training models or classifiers on different state-of-the-art datasets. Our goal is to bring these issues to the attention of the vision community. For this goal we propose in this paper a definition for measuring the training value of an example, and use this definition for ranking and greedily sorting examples.  After testing our methods on different vision tasks, models, datasets and classifiers, our experiments show that the performance of current state-of-the-art detectors and classifiers can be improved when training on a subset, rather than the whole training set.

\section{Related work}

Although training with the full training set is common practice, there are a few notable exceptions. Angelova et al. \cite{Angelova05} propose a method of pruning a training set and show its effectiveness in removing outliers and hard-to-learn examples. Within the context of a face recognition task, they introduce outliers (background samples labeled as faces) to the training set to test their algorithm. They train multiple classifiers using different partitions of the training set and identify which examples create more disagreement across classifiers, labeling them as troublesome. In our experiments we do not introduce any artificial noise because we can show that current datasets contain a large number of hard-to-learn examples for current recognition models, and prunning them can be beneficial. 

Some past approaches have tried to select training examples based on their effectiveness in training a model. The most common setting is active learning, whereby most of the data is unlabeled and an algorithm selects which training examples to label at each step, for the highest gains in performance. 
Thus, a principled ordering of examples can reduce the cost of labeling and lead to faster increases in performance as a function of amount of data available~\cite{GraumanIJCV2011}. Some active learning approaches focus on learning the hardest examples first (those closest to the decision boundary). 
Curriculum learning~\cite{BengioICML2009} instead advocates the opposite strategy, learning first from easy examples. 
Curriculum learning, however, requires a manual ranking of examples, and it is not always clear how to obtain it.

Robust learning algorithms provide an alternative way of differentially treating training examples, by assigning different weights to different training examples or by learning to ignore outliers~\cite{deICCV2001}. For example, Felzenszwalb et al.~\cite{FelzenPAMI2010} have shown that treating bounding boxes as noisy observations can provide significant improvements to the performance of the final classifier.  

Some methods for pruning training examples are designed to work with specific learning techniques. For boosting, Vezhnevets and Barinova \cite{vezhnevets2007avoiding} proposed an algorithm that eliminates training examples that are not correctly classified during the training, considering them as confusing examples. They showed their approach to be effective for improving classification performance when the classes have a large degree of overlap.  For SVMs, the loss function can be changed to become more robust to outliers and bad training examples \cite{wu07,brooks11}.

More recently, Zhu et al.~\cite{ZhuBMVC12} found that off-the-shelf detectors trained with a ``clean" subset of the training set may achieve higher performance than the full set. Similarly Razavi et al.~\cite{RazaviECCV2012} showed that excluding small clusters from the training set could improve performance. 
However, how to find the best ``clean" subset is an open problem. 
Pavlopoulou and Yu \cite{Pavlopoulou10} used human performance as a metric for weighting examples in terms of difficulty to improve performance on a scene recognition task.

Robust learning algorithms are designed to remove noisy examples. However, this is different from the problem we address here. We work with current vision datasets, which are highly curated and contain a reduced amount of noise. Our goal in this paper is to show that, in the context of a fixed model, some examples can be more valuable than others when learning a new concept. In this paper we argue that the most important issue is determining how to differentiate training examples into those that will help from those that will confuse the classifier. 

\section{Training Value of an Example}
\label{sec:TrainingValue}
 
\begin{figure}
\centering
\includegraphics[width=1\linewidth]{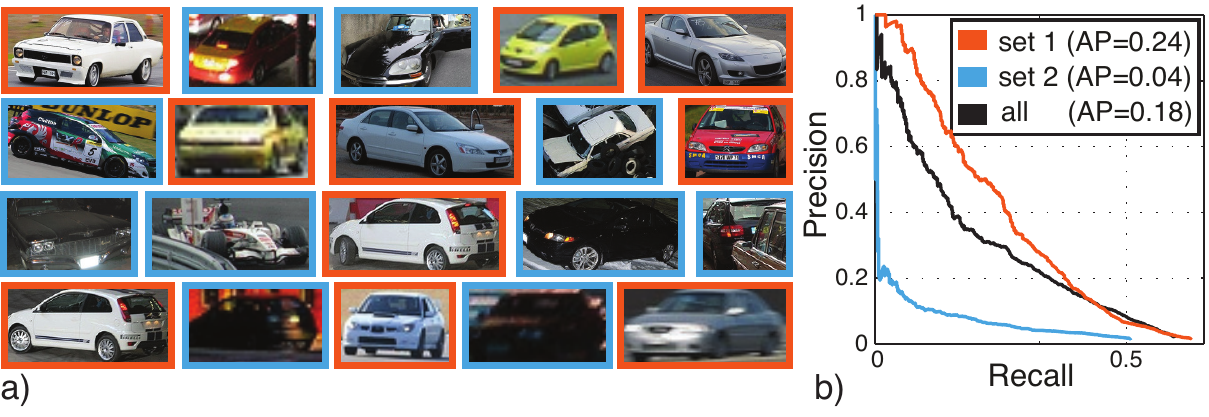}
\caption{Different performance for different subsets of training data. (a) 20 cars subsampled from PASCAL 2007 train+validation sets. When we train with all the instances in (a) and test on the PASCAL 2007 test set, we achieve the black PR curve in (b). However, when we train only on the cars in orange, instead of the full training set, we get the higher corresponding orange curve. Testing with the cars in blue produces the worst PR curve (in blue).}
\label{fig:multiplots}
\vspace{-.1in}
\end{figure}
 
Not all training examples are equal. A toy experiment in Fig.~\ref{fig:multiplots} demonstrates some of the points we will make throughout this paper. Consider the small dataset of 20 cars subsampled from the train+validation PASCAL 2007 ~\cite{PASCAL} of Fig.~\ref{fig:multiplots}\textcolor{red}{.a} for training a car detector using linear SVM on HOG features. Following standard protocol, we train a detector on this full dataset and test on the PASCAL 2007 test images containing cars to get the curve in black (Fig.~\ref{fig:multiplots}\textcolor{red}{.b}). Notice, however, that if we train only on the cars demarcated by orange boxes, then we obtain the orange curve with higher peformance! Alternatively, if we train on the cars demarcated by blue boxes, we get the curve with the lowest performance in blue. Thus, detectors we learn with different subsets of the training data vary in performance, and certain subsets can produce a detector that performs better than if trained on all the data. Notice that the cars in orange are more prototypical of their class, and it seems more intuitive that one would be able to learn a better model with them. But how can we directly measure the value of an example for training a detector, and further use this measure to rank training examples?
 
\begin{figure}
\centering
\includegraphics[width=1\linewidth]{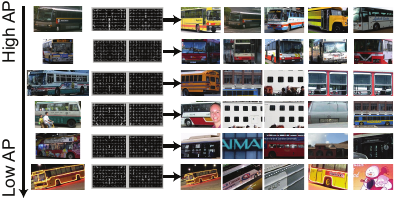}
\caption{Each row shows one example from the bus category, the HOG templates of an LDA-based exemplar detector trained with that example and then the 5 windows with the highest detection scores from the training set, under those models. Examples are organized from top to bottom according to the AP for detecting buses. The top two examples correspond to the examples that produce the best detectors, while the rest of examples are sampled uniformly from best to worst.}
\label{fig:SM2_bus}
\vspace{-0.2in}
\end{figure}

Given a particular task, a way of measuring the value of a training example for this task is treating it as a separate classifier. As a demonstration, we trained a separate LDA model for each example in the train+validation set of the PASCAL 2007 bus category, and evaluated AP on the same dataset. This is similar to the concept of Exemplar-SVMs~\cite{MalisICCV2011}, but where we use the performance of each exemplar detector to assign a training value to each example. Examples with high scores are more likely to be representative of the positive examples in the training set. The first column in Fig.~\ref{fig:SM2_bus} shows 5 training examples that were each used to train an LDA-based exemplar detector. The next column depicts the models (positive and negative) for each detector, followed by the top 5 detections on the training set under the models. The top 2 rows in the figure correspond to the training examples that produce the best detectors when evaluated on the training set. The next 3 rows correspond to training examples sampled uniformly from best to worst. Notice that the top 2 examples produce the best detectors when evaluated on the training set, while the last ones do not generalize as well. We use this idea to formally define the \emph{training value} of an example.

\subsection{Definition of Training Value}
\label{subsec:trainingValueDef}

Consider a binary classification task $\emph{\bf{T}}$ and a training dataset ${\bf X} = \{({\bf x_1},y_1),\dots,({\bf x_N},y_N)\}$, ${\bf x_i} \in \mathbb{R}^D$, $y_i \in \{-1,+1\}$, for $i=1,\dots,N$.
Let $\{f^{(\omega)},\omega \in \Omega\}$ be the set of a particular type of classification functions (for instance linear classifiers). Let ${\bf x_i}$ be a positive training sample. For the task $\emph{{\bf T}}$ we define the \emph{training value} of ${\bf x_i}$ according to the training set ${\bf X}$, as the average performance obtained by $\hat{f}_{{\bf x_i}}$ over the training set ${\bf X}$, where  $\hat{f}^{(\omega)}_{{\bf x_i}}$ is the classifier learned with the example ${\bf x_i}$ and all the negative examples in $\bf{X}$. Thus, we measure the absolute training value of  ${\bf x_i}$ as $AP(\hat{f}^{(\omega)}_{{\bf x_i}},{\bf X})$.

\section{Selecting training examples}

Just as there are some features that are more relevant than others for learning, there are some instances that are more informative than others. In this section we discuss the relationship between subset selection and feature selection and propose strategies for subset selection inspired by algorithms for feature selection. The experiments presented in the next section demonstrate the proposed methodologies to produce subsamplings of the whole training data that show improved performance of detectors and classifiers over training with the full training set.

\subsection{Feature versus example selection}

\begin{figure*}
\centering
\includegraphics[width=1.15\linewidth]{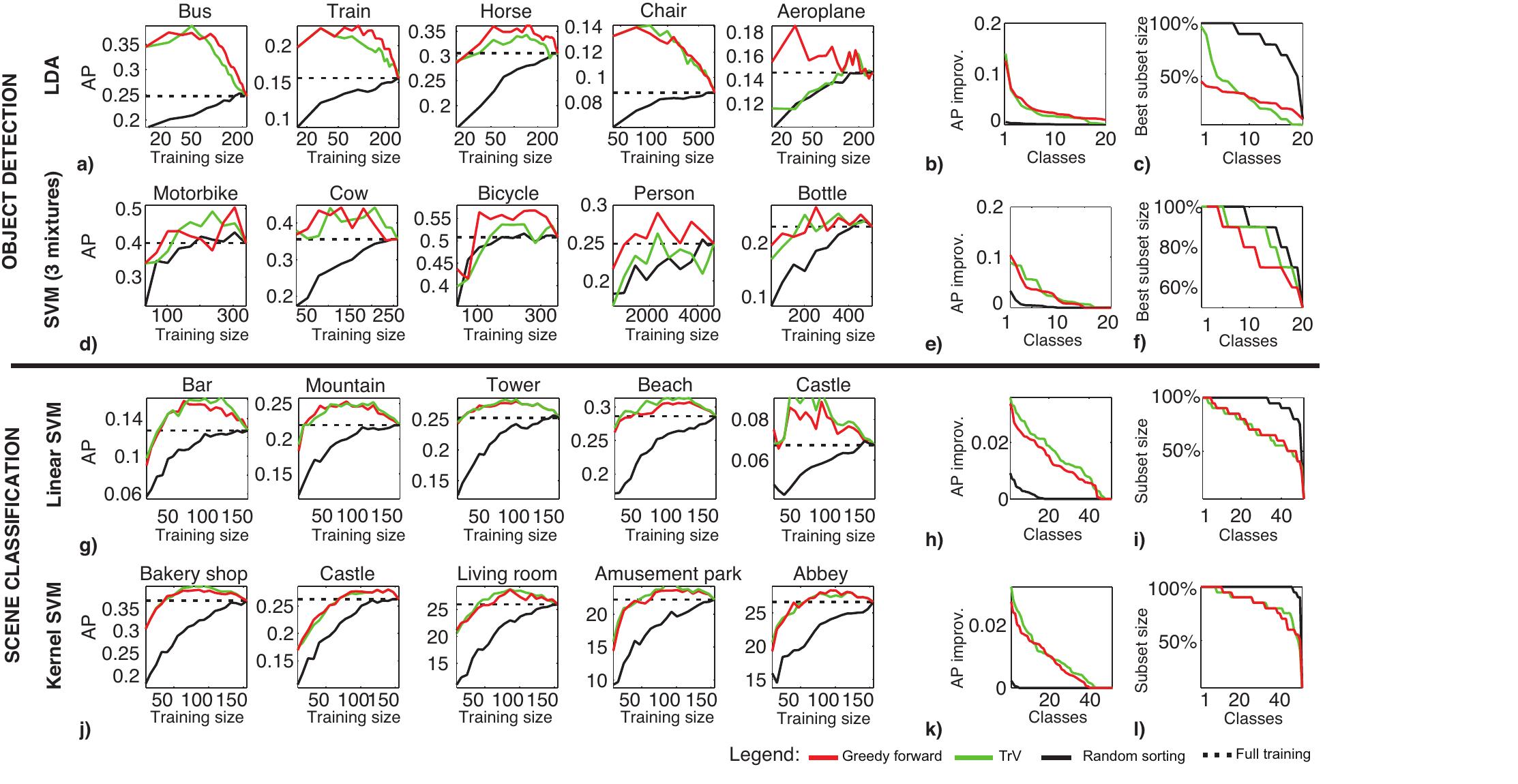}
\caption{We sort training data according to 3 methods (greedy, training value, and random), and evaluate the performance of a model using incrementally larger subsets of the data. Plotted are the 5 classes with largest improvements in AP using greedy to obtain a subset of the training data for testing (a) object detection with LDA, (d) object detection with linear SVM (3 mixtures), (g) scene classification with linear SVM, and (j) scene classification with kernel SVM. To summarize the rest of the classes, plotted is the absolute improvement in performance (in decreasing order) achieved on (b,e) all 20 PASCAL 2007 classes and (h,k) 50 SUN scenes. We also see which portion of the training set is used to reach a peak in performance for (c,f) object detection and (i,l) scene classification.}
\label{fig:mainResults}
\end{figure*} 

There is a large body of work on feature selection~\cite{guyon2003introduction, blum1997selection} for classification and prediction. The focus is usually on improving the accuracy and running time of the classifier as well as providing a better understanding of the data by selecting the best subset of features. Since we are interested in selecting the best subset of examples, our problem can be seen as the dual of the feature selection problem. For instance, in the case of a linear SVM, the decision function for a given data point ${\bf x_i}$ is $f({\bf x_i}) = {\bf w}^T{\bf x_i} = \alpha^T {\bf X} {\bf x_i}$ where rows of ${\bf X}$ are the training examples and ${\bf \alpha}$ is a vector of weights corresponding to the examples (with non-zero values for support vectors). In feature selection, we remove the non-valuable columns of ${\bf X}$ (equivalent to reducing the dimension of ${\bf x_i}$) while in example selection, we are interested in removing the non-valuable rows of ${\bf X}$ (equivalent to reducing the dimension of $\alpha$.) This relationship motivates adopting feature selection algorithms to example selection. 

Most feature selection approaches can be classified into {\em embedded}, {\em filter}, and {\em wrapper} methods~\cite{guyon2003introduction}. {\em Embedded} methods perform feature selection during training and are thus constrained by the model and learning algorithm. {\em Filter} methods evaluate each feature independently in order to choose the best subset. This approach is simple and has the benefit of being independent of the model and the training algorithm; however, it does not enforce diversity in the feature set. {\em Wrapper} methods utilize the learning algorithm as a black box and score the subsets of features based on their predictive or discriminative score. The subset of features can then be greedily grown (via greedy forward approaches) or shrunk (via greedy backward approaches). 

\subsection{Ranking and Greedy Sorting of Examples}
\label{sec:methods}

According to the definition of training value introduced in section \ref{subsec:trainingValueDef} we obtain a ranking of examples. This type of sorting can be seen as a dual version of a filter method for feature selection. 

Starting from this sorting, we propose a greedy forward methodology for selecting training examples. This greedy algorithm can be seen as the dual of wrapper methods in feature selection. In this case, we grow a subset of valuable examples by assessing the power of the model trained with them. This method, which treats the detector or classifier as a black box, provides a general tool that can be used in many situations. The main disadvantage of the greedy approach is the computational cost. However, our goal in this paper is to demonstrate the improvement in performance that can be gained with dataset selection, irrespective of the subsampling algorithm used. The details of the greedy strategy can be found in Algorithm {\color{red}1}.

\begin{algorithm}
\caption{Greedy sorting of training instances}
\label{alg:greedy}
\begin{algorithmic}
\STATE \textbf{Input:} \text{full training set}
\STATE \textbf{Output:} \text{greedy sorting $S$ of training set}
\STATE \textbf{Initialization:}
\STATE \text{(i) Sort training examples in decreasing order} 
\STATE \text{according to training value (as defined in section~\ref{subsec:trainingValueDef})}
\STATE \text{(ii) Split sorted training set into batches $B_i$ of equal size}
\STATE \text{(iii) $S \gets \{\}$}
\WHILE{\text{batches remain}}
\FORALL{\textrm{remaining batches $B_i$}}
\STATE {\text train on $S \cup B_i$ and evaluate on entire training set}
\ENDFOR
\STATE $S \gets S \cup B_j$ \textrm{ where $B_j$ is the batch that increases performance on the training set the most}
\ENDWHILE
\RETURN $S$
\end{algorithmic}
\end{algorithm}

\section{Experiments}

We use the methodologies described in section~\ref{sec:methods} to show that we can achieve gains in performance by taking subsets of the  whole dataset. Our experiments are carried out on object detection and scene classification tasks. We test on state-of-the-art datasets (PASCAL 2007 and SUN Database), features (HOG, Gist and Visual Words), and classifiers (linear SVM, kernel SVM, and LDA).

\subsection{Results on object detection}
\label{sec:results}
 
We perform object detection experiments using LDA and SVM (with three mixtures) on HOG features with the 20 object classes of VOC PASCAL 2007. To train we use the training+validation set and test on the testing set\footnote{For computational reasons we always test the detectos on all the images from PASCAL 2007 test dataset that contain instances of the tested classes}. For detection we use the frameworks of \cite{FelzenPAMI2010} and \cite{BharaECCV2012} respectively. 

Fig.~\ref{fig:mainResults}\textcolor{red}{.a-f} summarizes our results on object detection. Fig.~\ref{fig:mainResults}\textcolor{red}{.a-c} shows the results obtained with LDA while Fig.~\ref{fig:mainResults}\textcolor{red}{.d-f} those obtained with SVM (3 mix.). Fig.~\ref{fig:mainResults}\textcolor{red}{.a},\ref{fig:mainResults}\textcolor{red}{.c} show how AP on the test set changes when we train on more and more data. We incrementally add batches of data, sorted by decreasing training value (as defined in section \ref{subsec:trainingValueDef}). This is plotted in green. Furthermore we also plot in red the results obtained by sorting the data using the greedy method (described in section \ref{sec:methods}). As a baseline, we include a random sorting of the examples in black (averaging over 10 random sortings to enforce smoothness). We can see that a peak in performance is achieved before all the training data has been added. In fact, adding more data past a certain point actually causes performance to drop. We plot the 5 classes that show the most improvement on AP over using the entire training set. Fig.~\ref{fig:mainResults}\textcolor{red}{.b},\ref{fig:mainResults}\textcolor{red}{.e} summarize the absolute improvement in performance for all 20 classes, sorted in decreasing order of improvement. In Fig.~\ref{fig:mainResults}\textcolor{red}{.c},\ref{fig:mainResults}\textcolor{red}{.f} we plot the percent of training data that was used to reach the maximum improvement in performance for each of the classes. 

Observe that in the random sorting, the peak is usually not reached until all training instances have been added. However, when more informative sortings for the training instances are used, we see a peak in performance obtained with a subset of the whole data. Notice that for LDA (Fig.~\ref{fig:mainResults}\textcolor{red}{.c}) this peak always occurs when using less than 50$\%$ of the data. However, for SVM the size of the subsample is always larger than 50$\%$ (Fig.~\ref{fig:mainResults}\textcolor{red}{.f}). This suggests that the more robust the model, the more able it is to handle difficult data. The fact that the performance drops with more data past the peak (Fig.~\ref{fig:mainResults}\textcolor{red}{.a},\ref{fig:mainResults}\textcolor{red}{.d}) implies that some data can actually hurt the learning of the detector. But the more complex the model, the harder it is to hurt the detector.

Overall we obtained an improvement in AP on all 20 PASCAL 2007 classes with LDA and in 16 of them with SVM with a subset of the whole dataset. Thus we can learn better models if we don't use the full training data. 

\begin{figure}
\centering
\includegraphics[width=1\linewidth]{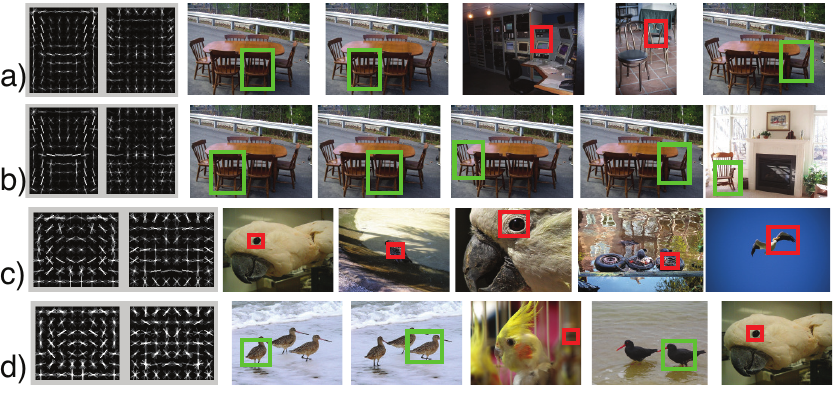}
\caption{Most confident (a) chair and (c) bird detections on the PASCAL 2007 test set for a detector trained with (a,c) the full training set from PASCAL 2007, (b,d) the subset selected by the greedy forward method: correct detections (green); false alarms (red).}
\label{fig:chair_detections}
\vspace{-0.2in}
\end{figure}

\begin{figure}
\centering
\includegraphics[width=1\linewidth]{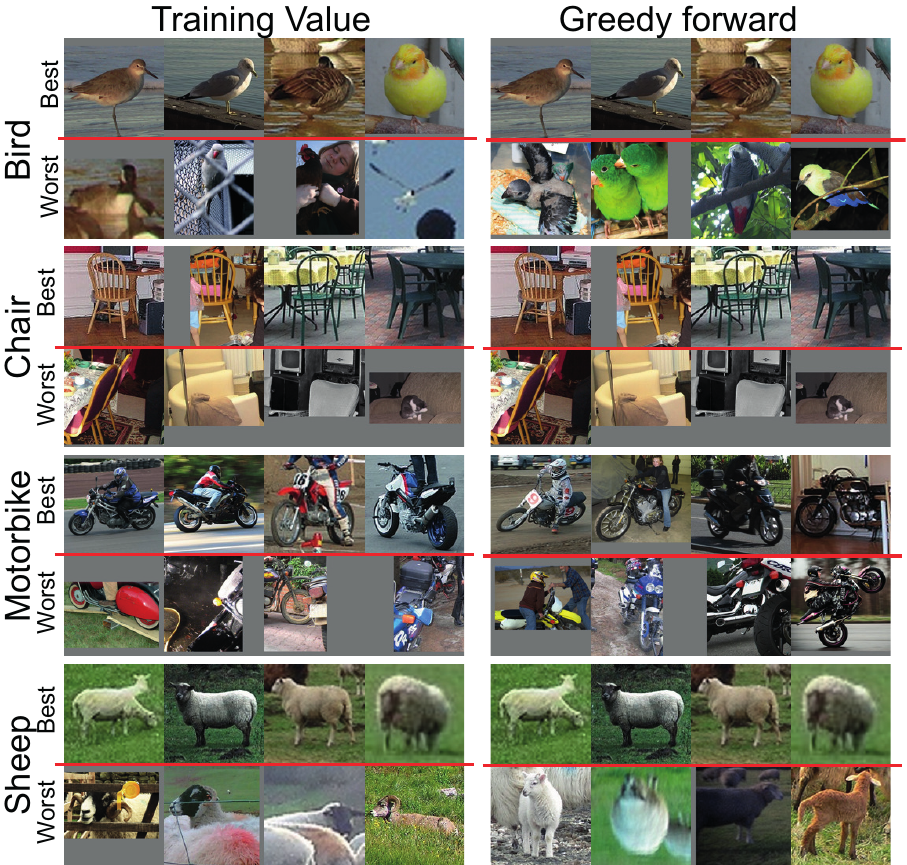}
\caption{For each panel of 8 examples, the top four correspond to the best examples and the bottom four are the worst examples. }
\label{fig:sortings}
\end{figure}

Fig.~\ref{fig:chair_detections} contains some qualitative results. The most confident detections obtained on the test set by training a model on the best subset (computed using the greedy method, Fig.~\ref{fig:chair_detections}\textcolor{red}{.b},\ref{fig:chair_detections}\textcolor{red}{.d}) for the {\em chair} and {\em bird} classes are much more intuitive (and in fact, more likely to be true positives) than the most confident detections obtained by training a model on the full training set (Fig.~\ref{fig:chair_detections}\textcolor{red}{.a},\ref{fig:chair_detections}\textcolor{red}{.c}).

\begin{figure}
\centering
\includegraphics[width=0.6\linewidth]{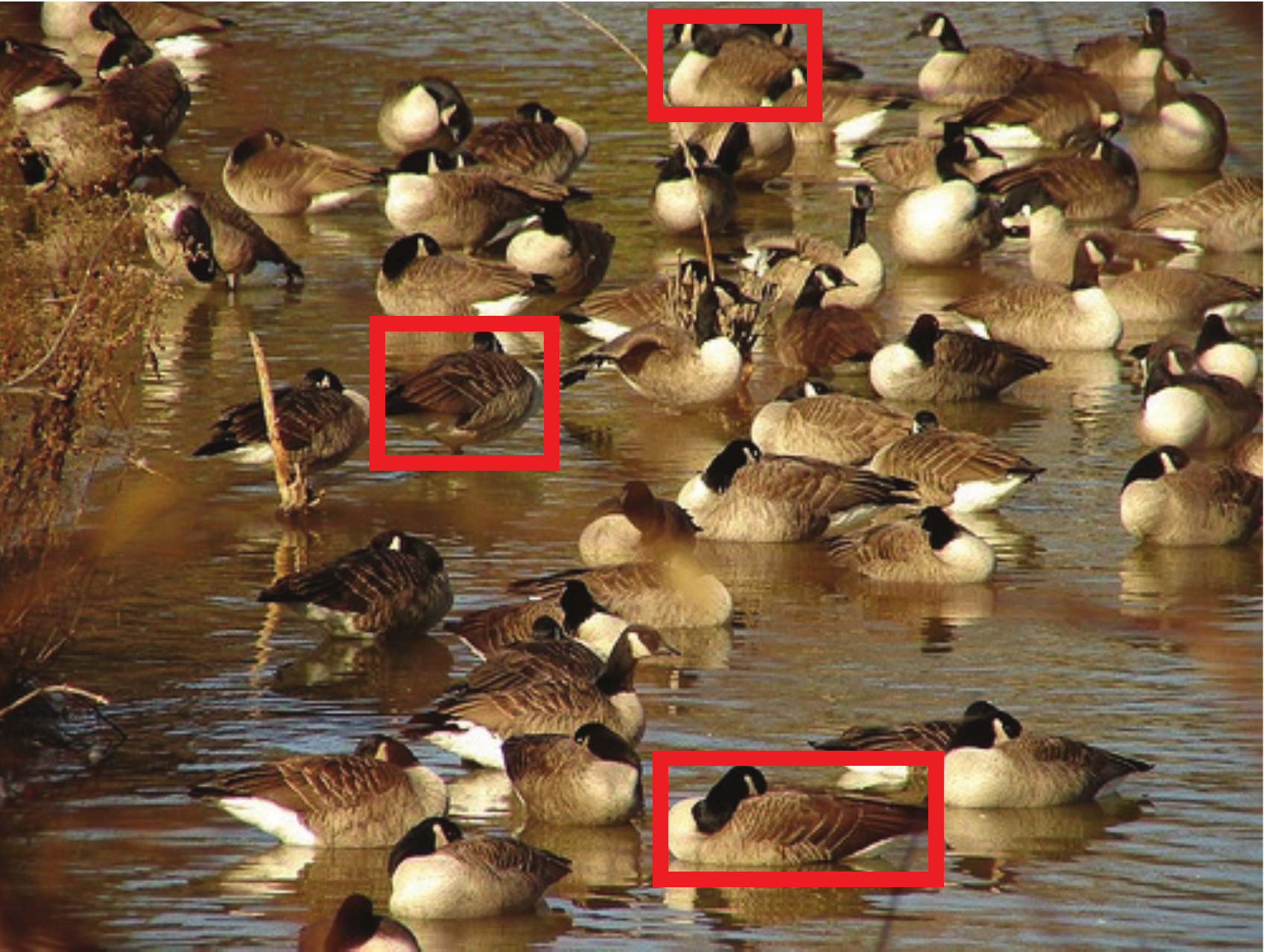}
\caption{Greedy chooses only 3 of these geese to train on.}
\label{fig:geese}
\end{figure}

In Fig.~\ref{fig:sortings}, we show the top and bottom 4 training examples for 5 classes, as ranked by training value and greedy forward selection methods, respectively. Because the ranking according to the training value evaluates the intances independently, the best instances are those that are most representative of their class. In particular, we can see that the top 4 training examples in Fig.~\ref{fig:sortings} tend to be prototypical, non-truncated, and non-occluded while the bottom 4 training examples tend to be highly occluded or truncated.  On the other hand, when we use a greedy forward selection approach, the distinction between high and low ranked examples is not so intuitive, because diversity enters into the picture. An example can be ranked low by the greedy method if it (a) can not be handled by the model; or (b) does not add diversity to the model. In the latter case, otherwise clean training instances may not be added because they would only add redundancy to the model. For instance, fig.~\ref{fig:geese} includes 30 of the 486 instances of birds present in the PASCAL training+validation set. Only three of these are included in the best subset obtained by the greedy method. That suggests that the others are likely to cause redundancy. This is a very natural setting: consider a dining hall of chairs, or a flock of birds. The likelihood that the instances will have similar appearance features and viewpoint is very high ({\em ``birds of a feather flock together"}).

\subsection{Results on scene recognition}

For the scene recognition experiments we used the 50 classes from the SUN database \cite{SUN} that had the most training examples of all SUN categories. From each image we extract visual words of HOG2x2 descriptors (the single best descriptor on the SUN database \cite{SUN}) and we build an image descriptor using the spatial pyramid \cite{Lazebnik06}. We train linear SVM and kernel SVM using histogram matching. This provides state of the art results on the SUN database. Here we train a classifier to discriminate one class (150 positive examples) versus 7750 negative examples (a random selection of images from all the other 900 SUN scene categories).

Fig.~\ref{fig:mainResults}\textcolor{red}{.g-l} shows the results on scene recognition. In order to apply the greedy method we evaluate each partition on a validation set to avoid the overfitting that can happen with kernel SVMs. The plots follow the same structure as in the case of object detection. Fig.~\ref{fig:mainResults}\textcolor{red}{.g},\ref{fig:mainResults}\textcolor{red}{.j} show AP as a function of the number of training examples (sorted by greedy method, training value and random sorting) for linear SVM and kernel SVM respectively. Again, we include curves for the five classes that show the most improvement in AP over using the entire training set. We also show in fig.~\ref{fig:mainResults}\textcolor{red}{.h},\ref{fig:mainResults}\textcolor{red}{.k} the absolute improvement in performance achieved per class, when sorting the classes in decreasing order. Fig.~\ref{fig:mainResults}\textcolor{red}{.i},\ref{fig:mainResults}\textcolor{red}{.l} depicts the percentage of examples included in the best subset. 

\begin{figure}
\centering
\includegraphics[width=1\linewidth]{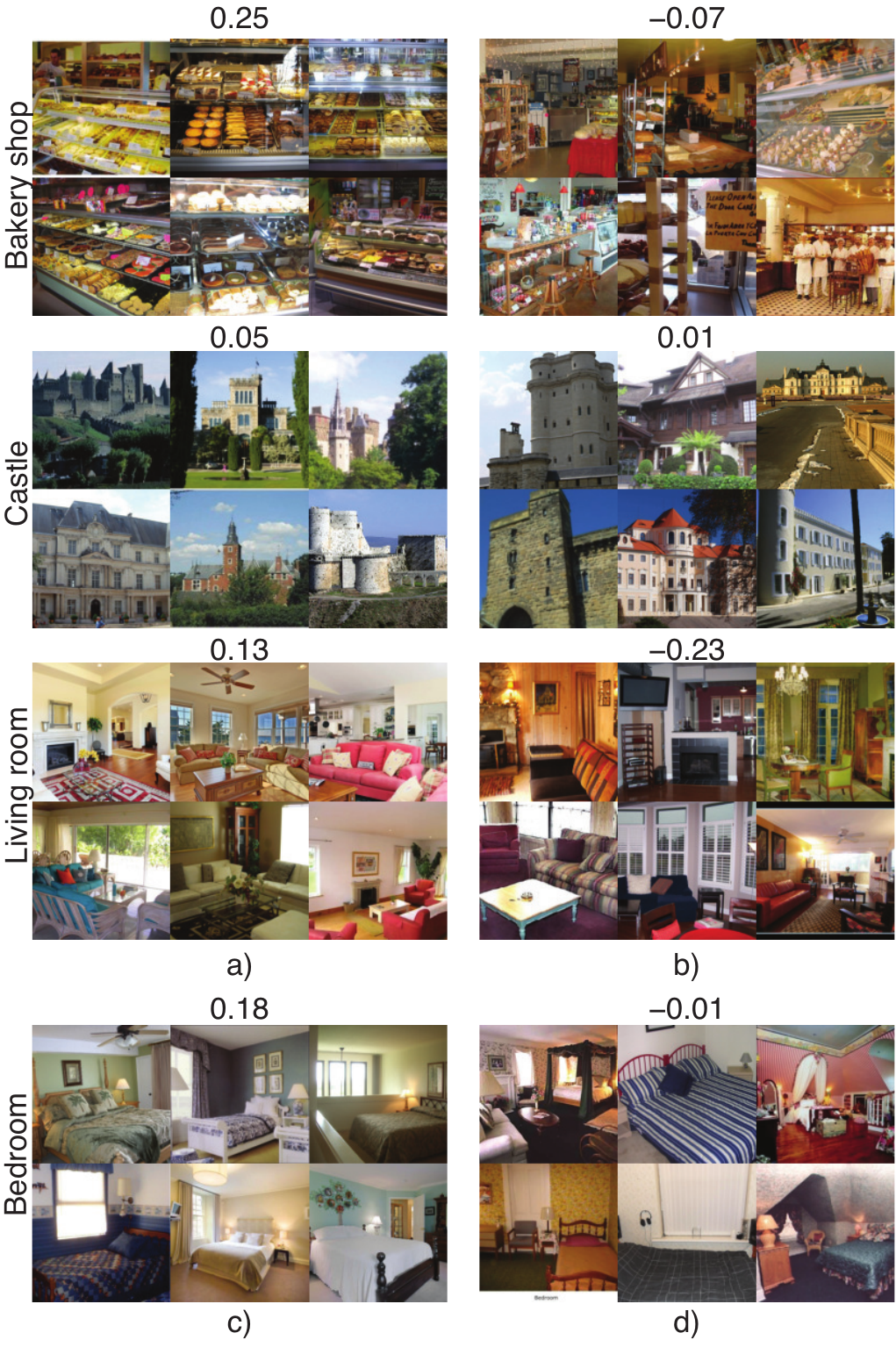}
\caption{Scene recognition. From the three scene categories with the strongest effect, we show (a) the 6 best, and (b) 6 worst examples selected by the greedy algorithm using a kernel SVM and HOG2x2. We also show  (c) the 6 best, and (d) 6 worst examples for the scene category that had the smallest improvement in performance. The number shown on top of each panel is the average (over 6 images) of the prototypicality index provided by humans from the SUN database (see section \ref{sec:prototypicality}). }
\label{fig:scene_recognition}
\end{figure}

Observe that, on average, we can find a subset of the training set that performs better than training on all the data. However, the differences are smaller than what is obtained in object detection for two main reasons:  a) we are using a stronger classifier able to learn from a larger variety of appearances, and b) scene databases are cleaner and contain, in proportion, fewer hard examples than object detection databases. Despite this, the selected examples (Fig.~\ref{fig:scene_recognition}\textcolor{red}{.a}) look more like prototypical members of the category than the worst examples, as chosen by the greedy ranking (Fig.~\ref{fig:scene_recognition}\textcolor{red}{.b}).  

\section{What makes a good example?}

In this section we want to address the question of whether particular attributes make some examples better than others, or if this is relative. We do show that for scenes, prototypicality (as rated by humans) correlates with training value. In the case of objects, truncated and occluded examples tend to be harder to learn from. In both cases, the visually cleanest examples are easier for classifiers and detectors. However, we go on to show that harder examples are not objectively bad. The difficulty of an example depends on the other examples available for training, and also on the model used and its complexity.

\subsection{Prototypicality}
\label{sec:prototypicality}

\begin{figure}
\centering
\includegraphics[width=1\linewidth]{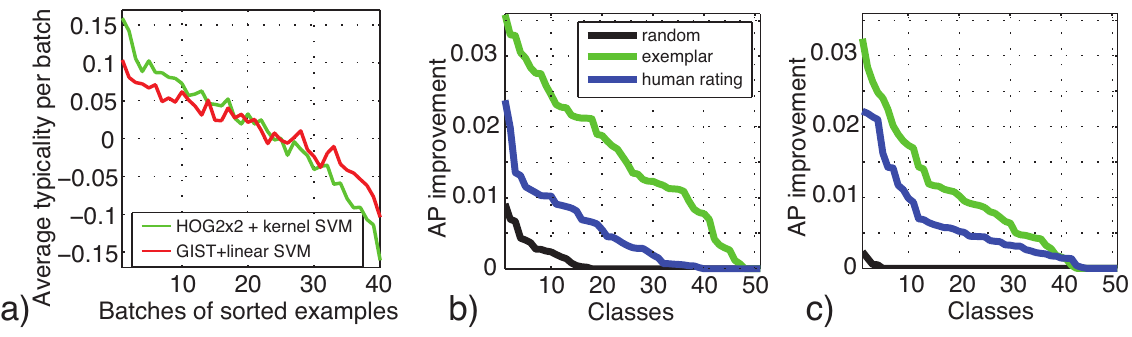}
\caption{Scene typicality. (a) The training examples that were assigned the highest typicality scores by humans occur in the initial batches of training data when sorted by training value. We also plot the absolute improvement in performance achieved per class by taking subsets of the training set sorted according to different metrics, where the training is performed using (b) HOG2x2 + kernel SVM, and (c) GIST + linear SVM. }
\label{fig:prototypicality}
\vspace{-.1in}
\end{figure}

From our qualitative results, our hypothesis was that prototypicality is a feature of instances with high training value. To test this hypothesis, we sorted the scene exemplars using human typicality rankings collected by Ehinger et al. \cite{Ehinger11}, and compared this sorting with the ones proposed in this paper. In fig.~\ref{fig:prototypicality}\textcolor{red}{.a} we plot the prototypicality of examples sorted according to their training value. We compute the training value for two models: HOG2x2 + kernel SVM (green curve) and GIST + linear SVM (red curve). The curves depict the mean prototypicality of sorted examples across the 50 classes of scenes used. In both cases we observe that, in general, examples with higher training value are rated more prototypical. In fig.~\ref{fig:prototypicality}\textcolor{red}{.b},\ref{fig:prototypicality}\textcolor{red}{.c} we can see the improvement in AP achieved for HOG2x2 + kernel SVM and GIST + linear SVM respectively, when sorting the samples randomly, by training value, and by prototypicality. In both plots, sorting by training value leads to larger gains in performance than sorting by prototypicality. However, this result is expected since training value is defined specifically to capture the quality of a training example as a classifier. The fact that sorting by prototypicality also leads to gains in performance seems to point to the fact that it is a useful feature for characterizing examples with high training value.

\subsection{Attributes}

Hoiem et al.~\cite{HoiemECCV12} performed an analysis of PASCAL bounding box annotations, including truncation, bounding box area and aspect ratio, as well as some other manual annotations. They focused on understanding where a detector fails. Here we use the same attributes but we look at a different problem: we are interested in understanding what makes a good {\em training} example.

We arrive at the conclusion that bounding box aspect ratio and viewpoint are only weak predictors of performance, and vary significantly from one class to the next. On the other hand, we observe that across classes, truncation is most correlated with the sorting of instances, when using training value for sorting. In particular, the number of truncated examples is higher, on average, in the later batches of data, than the ones added at the beginning. Occlusion and part visibility behave similarly. However, if we look at examples plotted by ranking order as specified by the greedy approach, the correlation with features is not as clean. In other words, image features are no longer as highly correlated with training value. This matches our previous observations, since the greedy approach also considers diversity an important feature for a training set to have. 

\subsection{Examples with low training value are better than nothing}

The fact that a significant portion of the training set makes the performance of a classifier drop does not mean that this portion of the dataset is composed of objectively poor or noisy training examples.

\begin{figure}
\centering
\includegraphics[width=1\linewidth]{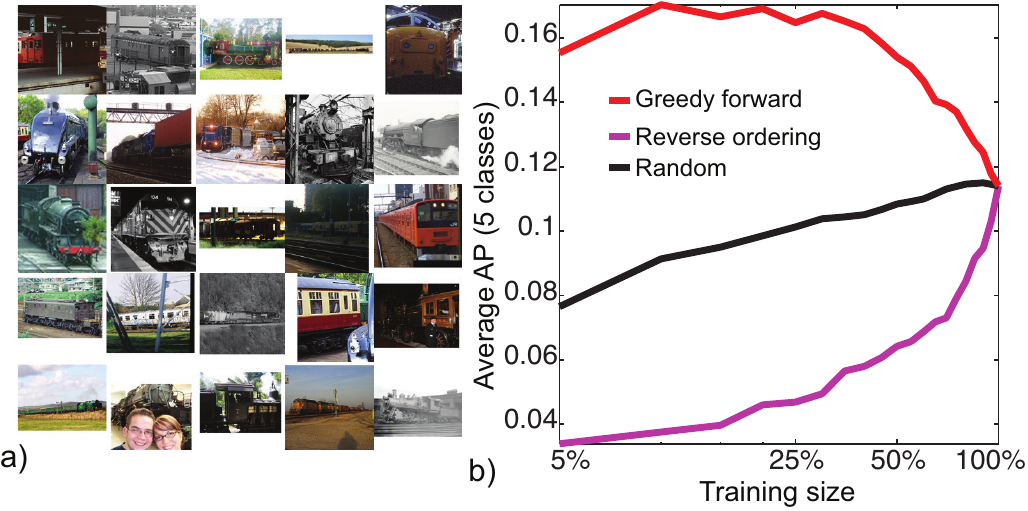}
\caption{Examples with low training value improve detector performance over chance level when no other training data is available. a) The 25 examples of the {\em train} class that were ranked last by the greedy forward method. b) Average AP on the test set over the 5 classes (bird, boat, bus, chair, and train) with the largest improvement (relative to the full training set). }
\label{fig:reverse}
\vspace{-0.1in}
\end{figure}

Whether or not examples affect a classifier negatively depends on the other examples available for training, as well as on the model used. As discussed in section \ref{sec:results}, the more complex the model, the harder it is to hurt it. Moreover, in the absence of better examples, examples that would otherwise be excluded can still be useful for training a detector.
To illustrate this, Fig.~\ref{fig:reverse} shows what happens if we reverse the ranking provided by the greedy forward method and we train the object detector with the examples with lowest training value first. The peformance is lower than what we achieve when training on the best subset, and even lower than using a subset of randomly chosen training examples. However, as we keep adding more training examples, the performance creeps up. Thus, it is not the case that these examples can not be used to learn a concept (they are not outliers, as we can see in Fig. \ref{fig:reverse}\textcolor{red}{.a}). We merely argue that they would not make for the most efficient nor the best way to learn the concept, and thus an informative sorting of examples can help us obtain higher performance with fewer training examples.

\section{Discussion}

\begin{figure}
\centering
a)\includegraphics[width=0.4\linewidth]{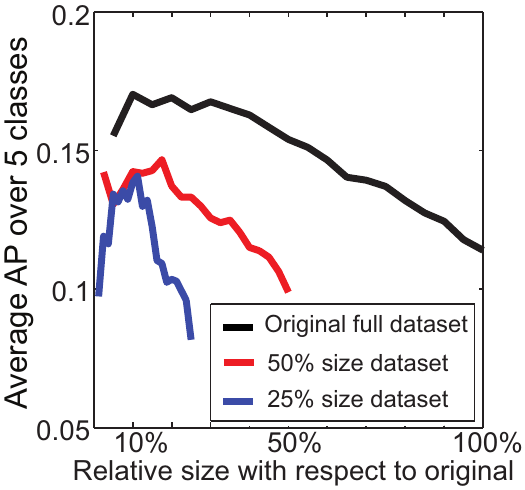}
b)\includegraphics[width=0.5\linewidth]{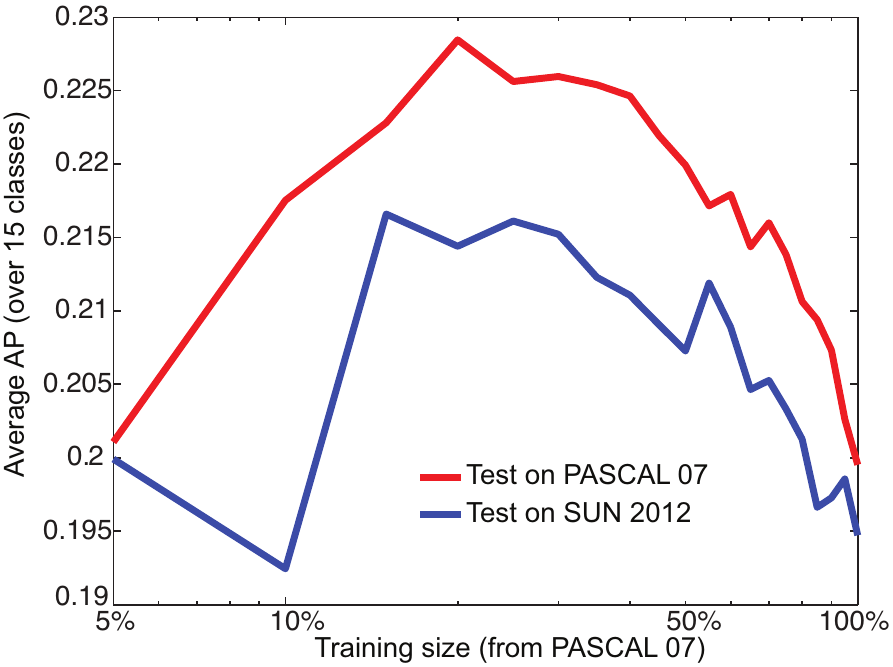}
\caption{a) Analysis of dataset size. The larger the dataset, the better the subset obtained from it. b) Analysis of dataset bias. The graphs show the average performance of 15 classifiers trained on PASCAL when tested on PASCAL and on SUN. }
\label{fig:datasetSize}
\vspace{-0.2in}
\end{figure}

{\bf Dataset size matters:} It is important to note that we do not want to conclude that using less data is better than using more data. In fact, when using a larger database the algorithm will be able to find a better subset of examples. Therefore, the overall performance is expected to grow as more data becomes available to choose from. Fig.~\ref{fig:datasetSize} illustrates this idea. We randomly selected 25$\%$ (blue) and 50$\%$ (red) of the original training set and ran our greedy forward on these new training sets. The plots are an average taken over 5 classes (bird, boat, bus, chair, train). For each training set size, we start with smaller batches of data and greedily add more data, re-training at each point, and evaluating on the test set. The trends are similar for all 3 curves. However, the more data we have to choose from, the higher the AP we can achieve on the test set.

{\bf Dataset bias:} One concern about the dataset selection method is that it overfits to the specific biases that one dataset might have (e.g., biases on view points, styles, etc.). As discussed in \cite{TorralCVPR2011} classifiers perform better when they are trained on the same dataset that they are tested on. Here we test what happens if we take the detectors trained on the PASCAL dataset and test them on the SUN dataset. If the selected subset has learned to generalize better then we should expect an increase in performance on another dataset with respect to a detector trained with the full dataset. This is what we observe in Fig.~\ref{fig:datasetSize}\textcolor{red}{.a}. In this experiment we use 15 classes that are common to the PASCAL and SUN datasets. Interestingly, the maximum performance (on average) is observed for the same subset from the PASCAL training set. This supports the idea that the classifier trained with a subset of the data generalizes better.

\section{Conclusion}

In this paper we show that some examples are better than others for training models or classifiers. We define a measure for the training value of an example, and propose methods for ranking examples based on this measure. We experimentally show that these methods are able to find subsets of the data that perform better than the whole training set in terms of AP on the testing set. In particular we show that we can improve with our approach the performance of a number of state-of-the-art classifiers and detectors (LDA, linear SVM, kernel SVM), and features (HOG, Gist, Visual Words) on popular vision datasets (PASCAL, SUN). We observe that some examples may negatively impact the performance of a classifier or a detector, and that removing them may be beneficial to training. Although standard practice is to use the entire dataset for training, we conclude from our study that sample selection is an important issue that should be taken into account by the general vision community.

{\small
\bibliographystyle{ieee}
\bibliography{egpaper_final}
}

\end{document}